\documentclass[conference]{IEEEtran}
\IEEEoverridecommandlockouts
\usepackage{cite}
\usepackage{amsmath,amssymb,amsfonts}
\usepackage{algorithmic}
\usepackage{graphicx}
\usepackage{textcomp}
\usepackage{xcolor}
\def\BibTeX{{\rm B\kern-.05em{\sc i\kern-.025em b}\kern-.08em
    T\kern-.1667em\lower.7ex\hbox{E}\kern-.125emX}}
\begin{document}

\title{Benchmarking CLIP for Zero-Shot Face and Periocular Gender Estimation\\
\thanks{Identify applicable funding agency here. If none, delete this.}
}

\author{
\IEEEauthorblockN{
Fernando Alonso-Fernandez$^{1}$,
Kevin Hernandez-Diaz$^{1}$,
Jose M. Buades$^{2}$,
Josef Bigun$^{1}$
}
\IEEEauthorblockA{
$^{1}$\textit{School of Information Technology},
Halmstad University, Sweden\\
$^{2}$\textit{Computer Graphics and Vision and AI Group},
University of Balearic Islands, Spain\\
feralo@hh.se, kevin.hernandez-diaz@hh.se,
josemaria.buades@uib.es, josef.bigun@hh.se
}
}

\maketitle

\begin{abstract}
We investigate CLIP for zero-shot gender estimation from full-face and periocular images. 
Three CLIP backbones are evaluated on 11,299 frontal images from Adience using image-text similarity with male/female prompts, 
achieving 95.54\% full-face accuracy without task-specific training.
For periocular, zero-shot predictions are strongly biased towards males, primarily due to a misaligned decision boundary. Threshold alignment substantially reduces this bias, reaching 85.29\% accuracy.
Linear SVMs trained on CLIP features provide only marginal gains, with a best periocular accuracy of 86.17\%, approximately 2.8\% above previous Adience results in the literature. 
Nevertheless, the gap with full-face performance confirms the greater difficulty of periocular gender estimation.
\end{abstract}


\begin{figure*}[h]
  \centering
  \includegraphics[width=0.85\textwidth]{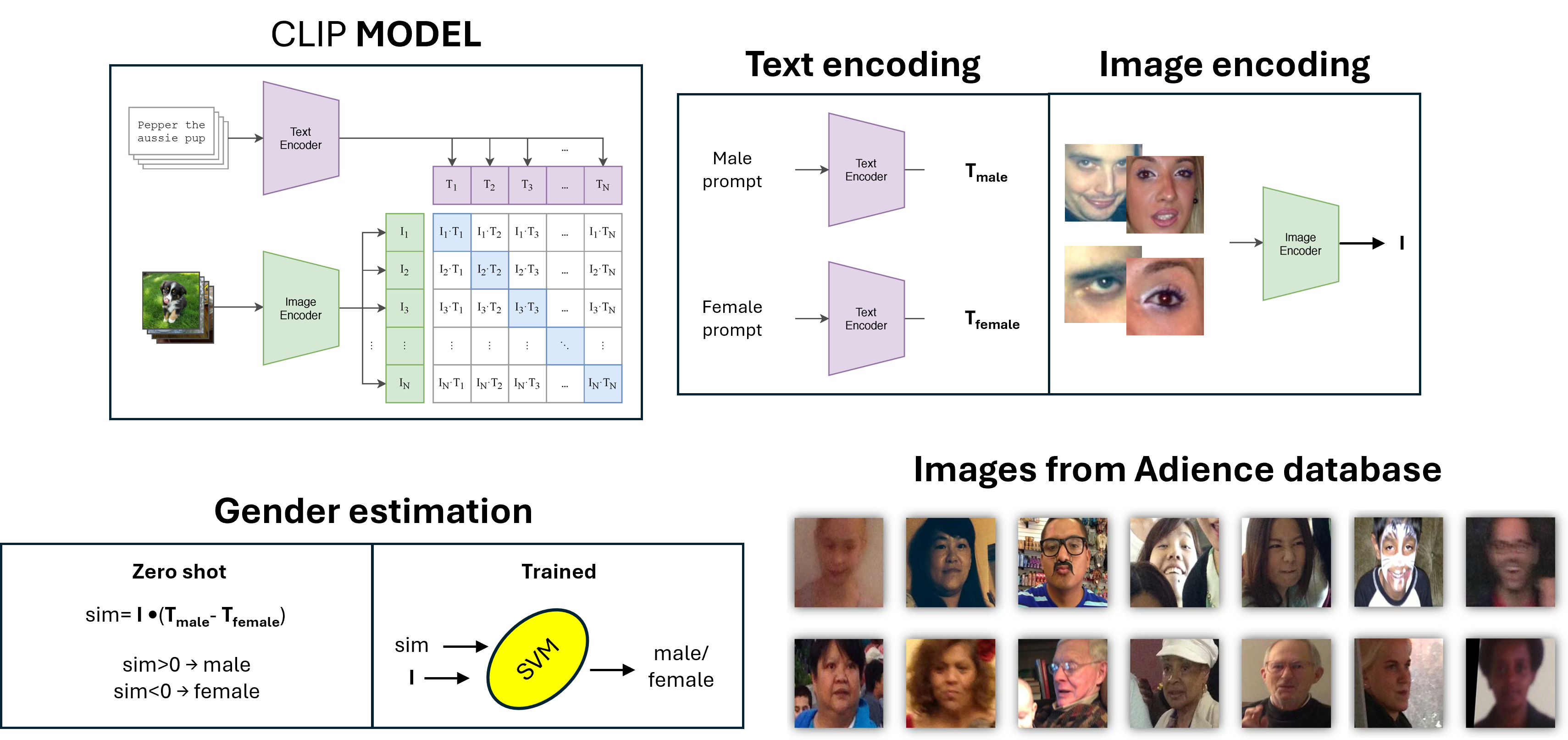}
  \caption{Top left: CLIP model architecture (from \cite{2021_RadfordICML_CLIP}). Top-right and bottom-left: our adaptation for face/periocular gender estimation. Bottom right: images from the Adience database (from \cite{[EidingerHassner14_Adience]}).}
  \label{fig:clip_framework}
\end{figure*}

\section{Introduction}
\label{sec:intro}
Soft biometrics refers to the estimation of semantic attributes such as gender, age, or ethnicity from biometric data \cite{[Dantcheva16softbio]}. 
Although these attributes are generally insufficient for identification on their own, they provide complementary information for biometric recognition and related applications such as subject filtering or attribute-dependent services \cite{Becerra2019airFaceSoftbioForensicSurveillSurvey}. 
Full-face images are widely used because they contain rich semantic information, but the full face is not always visible or reliable due to pose variations, illumination changes, partial occlusions, etc. \cite{2026PAMI_50yrfacerecognition_jain}.
%
%
This has motivated increasing interest in the periocular region, which often remains visible when other facial areas are occluded and contains useful information for recognition and soft-biometric estimation \cite{sharma23cviu_periocular_masks_survey,Alonso24computers_periSOA}.
Nevertheless, periocular analysis remains more challenging because it relies on a smaller facial region.

Gender estimation from face and periocular images has traditionally relied on hand-crafted descriptors or supervised deep learning, requiring labelled data or task-specific training. 
Foundation models, particularly vision-language models (VLMs), offer an alternative through large-scale pre-training and transfer with little or no task-specific supervision \cite{2025_OtroshiTIFS_Foundation_Models_Biometrics_Survey}. One of the pioneering VLMs, CLIP \cite{2021_RadfordICML_CLIP}, for example, learns a joint image-text embedding space that enables visual concepts to be queried with natural-language prompts.
Recent studies have explored VLMs for soft-attribute estimation, primarily from full faces. 
GPT-4V has been evaluated for face biometrics and soft attributes \cite{2024_HassanpourICIP_ChatGPTBiometrics,2024_DeAndresACCESS_ChatGPT_Face_Biometrics}, while broader benchmarks have assessed multiple VLMs and multimodal models for gender, age, ethnicity, and related face-analysis tasks \cite{2025_SonyRoss_Benchmarking_FMs_Biometrics,2026_NarayanPatelTBIOM_FaceXBench_LLMs}. 
Other works have adapted VLMs specifically for fine-grained facial understanding to counteract the lack of face image-text training data in general models, including Face-MLLM \cite{2024_Sun_FaceMLLM} and FaceInsight \cite{2025_LiACM_ICM_FaceInsight}.

Previous works mainly focus on full-face images or analyse the textual responses generated by the model.
In contrast, we investigate VLM-based gender estimation from both full-face and periocular images and explicitly exploit image-text similarity rather than relying only on visual embeddings or generated textual responses. 
We evaluate three CLIP backbones on Adience \cite{[EidingerHassner14_Adience]} using zero-shot male/female prompts. 
While CLIP transfers effectively to full-face gender estimation, periocular zero-shot predictions exhibit a strong male bias due to a misaligned decision boundary. A simple threshold alignment largely corrects this bias, while supervised SVMs trained on CLIP features provide only modest additional gains, suggesting that most of the gender-discriminative information is already captured by the CLIP embedding space. 
Our best accuracies reach 95.87\% for full-face and 86.17\% for periocular images, the latter outperforming previous Adience results by 2.8\%. To the best of our knowledge, this is the first study of VLMs for periocular gender estimation.

\begin{table*}[tb]
\centering
\caption{CLIP models evaluated in this paper.}
\resizebox{0.75\textwidth}{!}{%
\begin{tabular}{|cccccc|ccc|}

\multicolumn{6}{c}{Image encoder} & \multicolumn{3}{c}{Text encoder}  \\ \cline{1-9}

\multicolumn{1}{|c}{} & Input & Input &  &  Para- & Vector &  & Para- & Vector \\

\multicolumn{1}{|c}{\textbf{Model}}  & size & patches & Layers & meters & Size & Layers & meters & size  \\ \hline

ViT-B/16 & 224$\times$224 & 16$\times$16 & 107 & 86.2M  & 512 & 104 & 63.4M  & 512  \\

ViT-L/14 & 224$\times$224 & 14$\times$14 & 203 & 304M & 768 & 104 & 123.7M  & 768 \\

ResNet-50 & 384$\times$384 & - & 437 & 167.4M & 768  & 104 & 123.7M  & 768  \\ \hline

\end{tabular}
}
\label{tab:networks}
\end{table*}

\begin{table*}[!h]
\centering
\caption{Prompts employed for face/periocular gender classification.}
\resizebox{0.85\textwidth}{!}{%
\begin{tabular}{lll}
\hline


\multicolumn{1}{|l|}{\#} & \multicolumn{1}{l|}{\textbf{Face}} & \multicolumn{1}{l|}{\textbf{Periocular}} \\ \hline

\multicolumn{1}{|l|}{1} & \multicolumn{1}{l|}{a face photograph of a man/woman} & \multicolumn{1}{l|}{a photograph of a male/female eye} \\ \hline

\multicolumn{1}{|l|}{2} & \multicolumn{1}{l|}{a face image of a man/woman} & \multicolumn{1}{l|}{an image of a male/female eye} \\ \hline

\multicolumn{1}{|l|}{3} & \multicolumn{1}{l|}{a photograph of a man/woman} & \multicolumn{1}{l|}{an eye image of a man/woman} \\ \hline

\multicolumn{1}{|l|}{4} & \multicolumn{1}{l|}{a photograph of a male/female person} & \multicolumn{1}{l|}{an ocular image of a man/woman} \\ \hline

\multicolumn{1}{|l|}{5} & \multicolumn{1}{l|}{a photograph of a male/female human face} & \multicolumn{1}{l|}{an ocular image of a male/female eye} \\ \hline

\multicolumn{1}{|l|}{6} & \multicolumn{1}{l|}{a portrait photo of a man/woman} & \multicolumn{1}{l|}{a photograph of a male/female person's eye} \\ \hline

\multicolumn{1}{|l|}{7} & \multicolumn{1}{l|}{a portrait of a male/female person} & \multicolumn{1}{l|}{a detailed image of the eye of a male/female person} \\ \hline

\multicolumn{1}{|l|}{8} & \multicolumn{1}{l|}{a close-up photograph of a man's/woman's face} & \multicolumn{1}{l|}{a close-up photograph of a male/female human eye} \\ \hline

\multicolumn{1}{|l|}{9} & \multicolumn{1}{l|}{a close-up face image of a man/woman} & \multicolumn{1}{l|}{a close-up photo of a male/female human eye} \\ \hline

\multicolumn{1}{|l|}{10} & \multicolumn{1}{l|}{a close-up photo of a male/female face} & \multicolumn{1}{l|}{a close-up image of a male/female human eye} \\ \hline 

\\
\end{tabular}
}
\label{tab:prompts}
\end{table*}


\section{Method}
\label{sec:method}
CLIP (Contrastive Language-Image Pre-training) \cite{2021_RadfordICML_CLIP} is a VLM trained through contrastive learning on 400 million image-text pairs to predict which caption goes with which image (Figure~\ref{fig:clip_framework}, top left). 
It maps images and text into a common embedding space, enabling zero-shot transfer through natural-language queries to reference learned visual concepts. 
We evaluate three pre-trained CLIP image backbones: ViT-B/16, ViT-L/14, and a ResNet-50 with 16× width scaling. Their main characteristics are summarised in Table~\ref{tab:networks}.

For gender estimation, we define male/female prompts for full-face and periocular images (Table~\ref{tab:prompts}). Given a full face, CLIP produces an image embedding $\mathbf{I}$. For periocular images, embeddings are extracted independently from the left and right eyes, $\mathbf{I_L}$ and $\mathbf{I_R}$, and $\mathbf{I}$ is computed as their average. In parallel, the text encoder produces embeddings $\mathbf{T}_{\mathbf{male}}$ and $\mathbf{T}_{\mathbf{female}}$ for the corresponding prompts.
Gender similarity is then computed from the L2-normalised embeddings as $sim = \mathbf{I} \cdot (\mathbf{T}_{\mathbf{male}} - \mathbf{T}_{\mathbf{female}})$, which is equivalent to the difference in cosine similarity with the male and female text embeddings.
Images with $sim > 0$ are classified as male, and as female otherwise.
The complete zero-shot procedure is illustrated in Figure~\ref{fig:clip_framework} (bottom left). 
We additionally train linear SVM classifiers \cite{[Vapnik95]} using the CLIP similarity scores, image embeddings, or their combination as input features.

\begin{figure*}[!htb]
  \centering
  \includegraphics[width=0.75\textwidth]{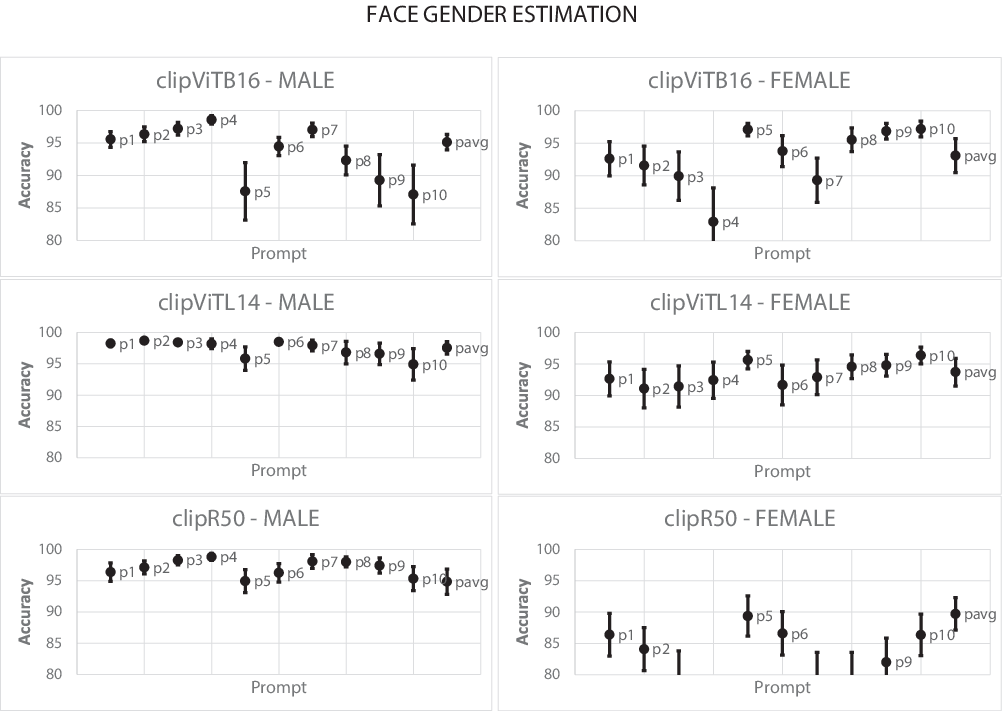}
  \includegraphics[width=0.55\textwidth]{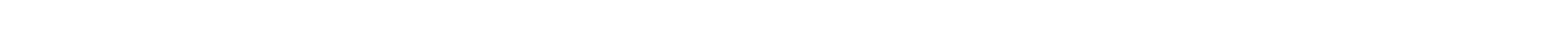}
  \includegraphics[width=0.75\textwidth]{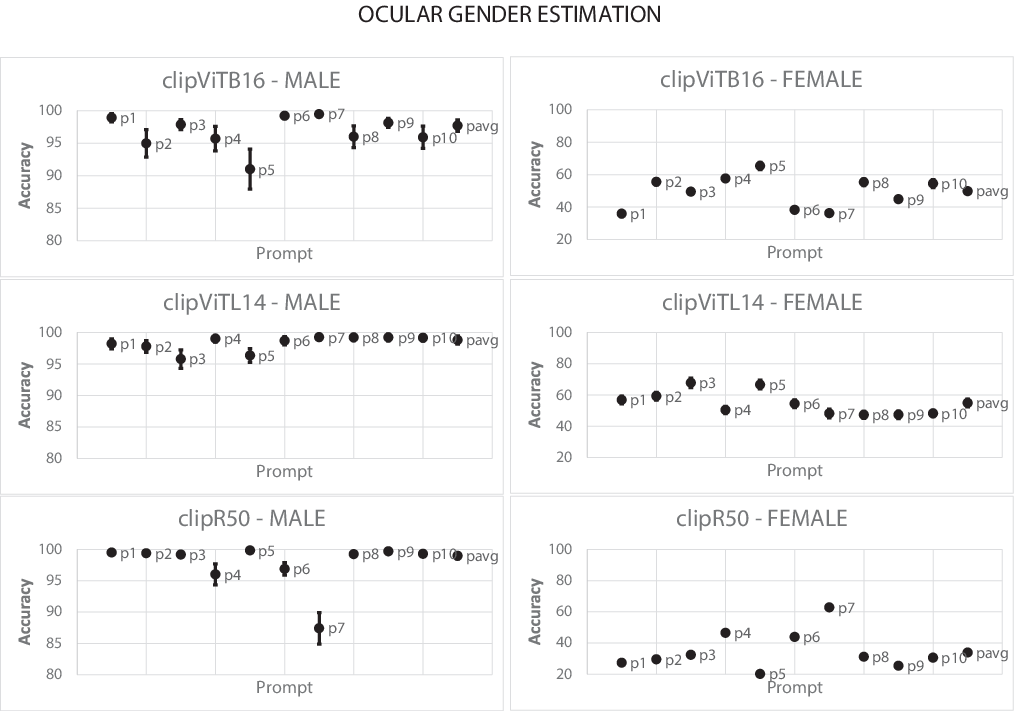}
  \caption{Zero-shot \textbf{face/periocular} gender estimation accuracy per prompt on the Adience database for different CLIP backbones. The dots indicate the mean accuracy over the five cross-validation test folds, while error bars represent one standard deviation.}
  \label{fig:adience_prompts_v1_accuracy_zero_shot}
\end{figure*}


\section{Database and Protocol}
\label{sec:exp_framework}

We use the Adience benchmark \cite{[EidingerHassner14_Adience]} with unconstrained Flickr images captured with smartphones (Figure~\ref{fig:clip_framework}, bottom-right). 
From the original 26,580 images of 2,284 subjects, we retain only 11,299 frontal images (5,353 males, 5,946 females) to simulate selfie captures. 
Face landmarks are detected with MTCNN \cite{[Zhang16]}, and images are aligned using the eye axis and normalised to an inter-eye distance of 105 pixels (database average). 
We then extract 224$\times$224 face crops centred on the landmarks' centroid and 113$\times$113 left/right periocular crops.
We follow the predefined 5-fold cross-validation protocol, which prevents Flickr album overlap between the training/test sets. 
Supervised models are trained on the remaining four folds, and results are reported as mean accuracy and standard error over the five folds. CLIP inputs are resized according to Table~\ref{tab:networks}.
As mentioned in Section~\ref{sec:method}, classification is performed either from CLIP image-text similarities (zero-shot) or with a linear SVM trained on similarity scores, image embeddings, or their combination. Experiments use pretrained CLIP models in MATLAB R2026a on an NVIDIA RTX 2080 Ti GPU.

\begin{figure*}[!h]
  \centering
  \includegraphics[width=0.75\textwidth]{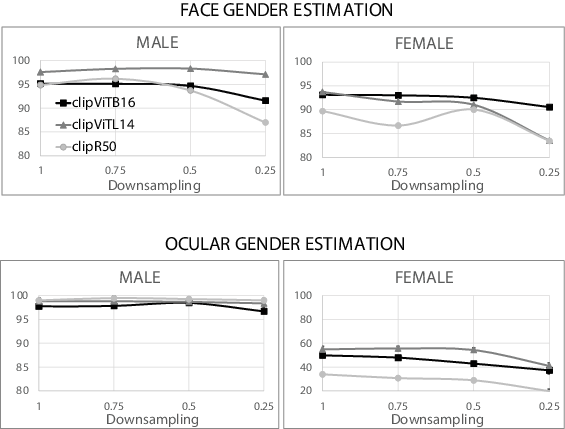} 
  \caption{Zero-shot \textbf{face/periocular} gender estimation accuracy on the Adience database for different CLIP backbones and downsampling values (mean accuracy over the five cross-validation test folds). Results correspond to the average encoding of all prompts. 
  }
  \label{fig:adience_face_ocular_scales_zero_shot}
\end{figure*}


\section{Results}
\label{sec:results}

\subsection{Zero-shot face and periocular gender estimation.}

Zero-shot gender estimation results are shown in Figure~\ref{fig:adience_prompts_v1_accuracy_zero_shot} for the ten prompts (p1, p2, etc.) and their averaged text embedding (pavg).
Numerical pavg results are reported in Table~\ref{tab:results_all} (``sim (zero shot)'' row).
Full-face estimation performs consistently well across all CLIP backbones, with per-class accuracies of many prompts above 90\%. 
In contrast, periocular performance shows a strong bias: male accuracy exceeds 95\% for most prompts, whereas female accuracy is often below 60\% and falls below 30\% for some ResNet-50 prompts.
Prompt choice affects performance, but no prompt is consistently optimal across classes and backbones. 
Averaging the prompt embeddings (pavg) provides intermediate, more stable performance rather than systematically outperforming individual prompts. 
With pavg, overall accuracy ranges from 92.17\% to 95.54\% for full-face images and from 64.59\% to 75.57\% for periocular images.
Among the three backbones, ViT-L/14 achieves the best results and ResNet-50 the worst. Despite having fewer parameters, ViT-B/16 outperforms ResNet-50, suggesting that transformer-based backbones may better capture gender-related information. The poorer ResNet-50 performance may also be influenced by its larger input resolution, which requires stronger upsampling, particularly for periocular crops.

\begin{figure*}[h]
  \centering
  \includegraphics[width=0.8\textwidth]{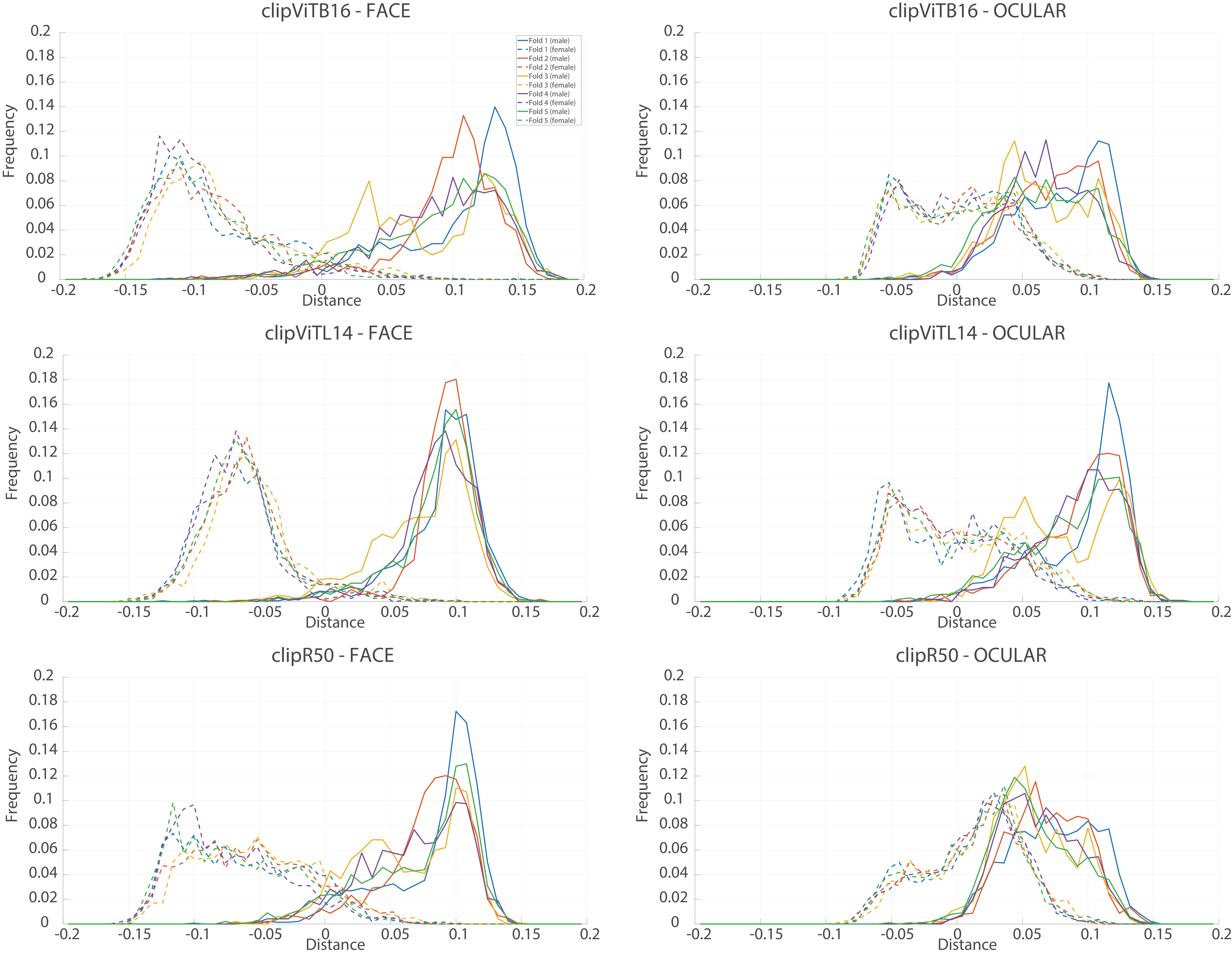}
  \caption{Histograms of zero-shot \textbf{face/periocular} scores on the test folds of the Adience database. 
  Results correspond to the average encoding of all prompts.}
  \label{fig:adience_prompts_v1_histograms_zero_shot}
\end{figure*}

\subsection{Scale analysis.}

To assess whether the poorer periocular performance is caused by lower spatial resolution, we downsample the input images before feeding them to CLIP (Figure~\ref{fig:adience_face_ocular_scales_zero_shot}). Downsampling by up to a factor of 2 (0.5) produces only minor accuracy reductions for both face and periocular, indicating robustness to moderate resolution loss.
Even at 0.25 scale, full-face female accuracy remains higher than periocular female accuracy at the original resolution, while male periocular accuracy remains above 96\%. These results indicate that limited spatial resolution alone cannot explain the poor zero-shot performance for female periocular samples, pointing instead to limitations in the visual representations learned by CLIP.

\begin{figure*}[!htb]
  \centering
  \includegraphics[width=0.75\textwidth]{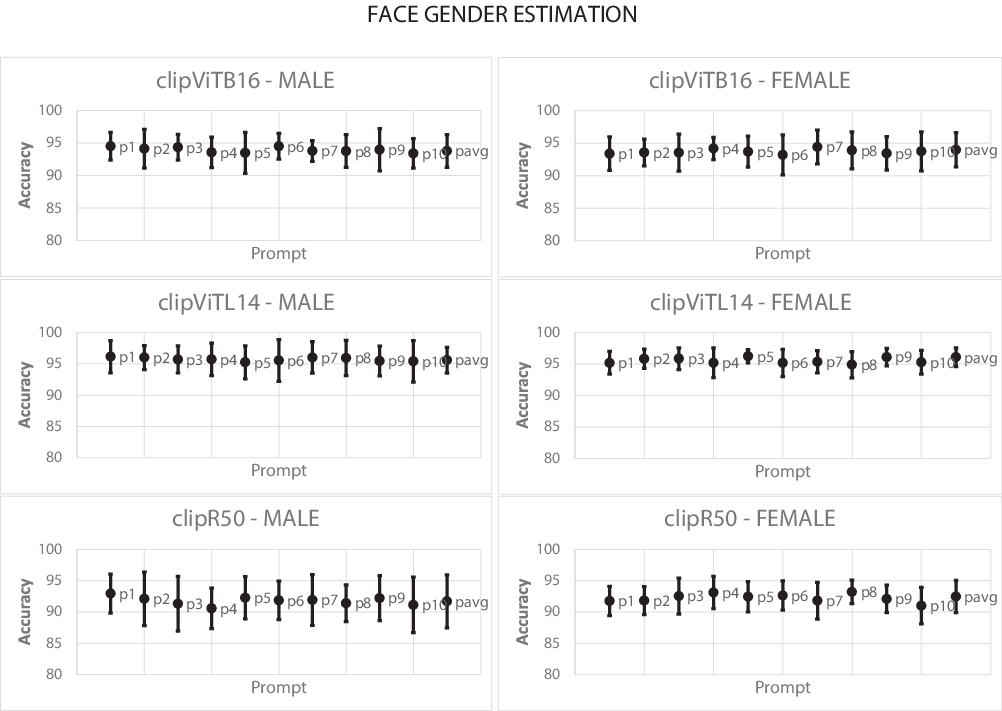}
  \includegraphics[width=0.55\textwidth]{BLANK.png}
  \includegraphics[width=0.75\textwidth]{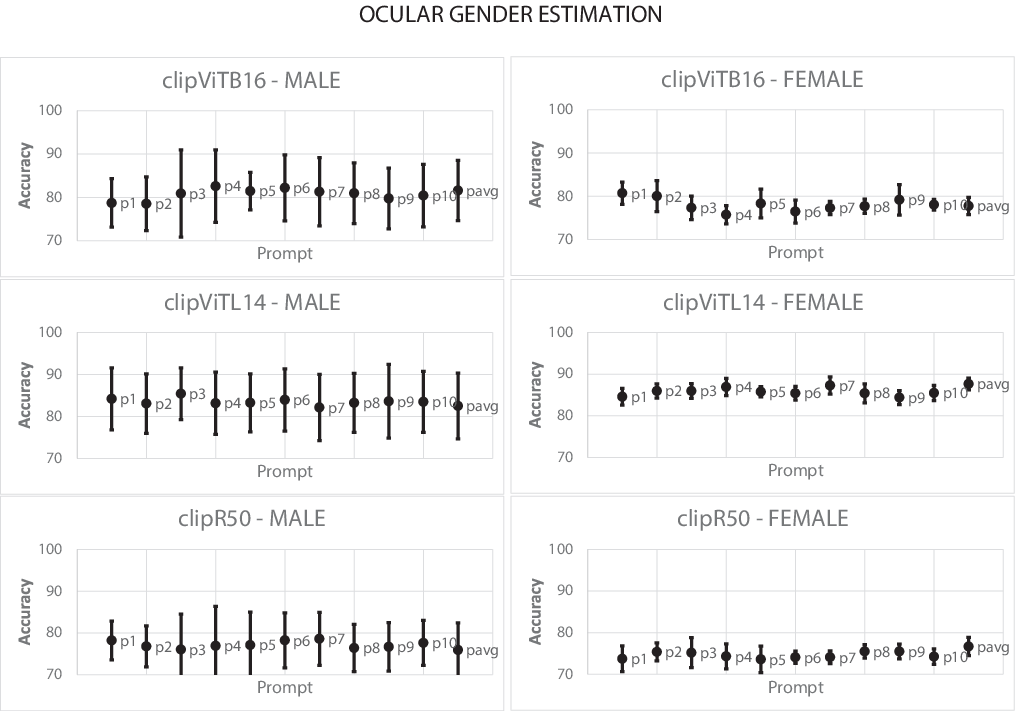}
  \caption{\textbf{Face/periocular} gender estimation accuracy per prompt \textbf{after embeddings alignment} on the Adience database for different CLIP backbones. The dots indicate the mean accuracy over the five cross-validation test folds, while error bars represent one standard deviation.}
  \label{fig:adience_prompts_v1_accuracy_zero_shot_aligned}
\end{figure*}

\subsection{Embeddings alignment.}

To analyse the male bias in periocular zero-shot predictions, we examine the similarity-score distributions using pavg in Figure~\ref{fig:adience_prompts_v1_histograms_zero_shot}. 
For full-face images, male and female scores are well separated around the zero threshold. In contrast, periocular female scores are shifted towards positive values, causing many female samples to be classified as male. The larger overlap between male and female periocular distributions also suggests that performance will remain below full-face estimation even after correcting this shift.
We therefore replace the zero threshold with a prompt-specific threshold estimated on the training folds by maximising balanced classification accuracy. 
This alignment substantially improves female periocular accuracy from 33.92\%-54.86\% to 76.67\%-87.64\%, while male accuracy decreases from 97.7\%-98.97\% to 75.93\%-82.5\%, resulting in a more balanced performance per prompt and per class (Figure~\ref{fig:adience_prompts_v1_accuracy_zero_shot_aligned}, Table~\ref{tab:results_all}). 
Full-face accuracy changes little because its scores are already well aligned around zero. Periocular performance nevertheless remains lower due to the greater overlap between the mentioned class distributions.

\begin{table*}[h]
\centering
\caption{Comparative summary of face/periocular gender classification on the Adience database with the different methods evaluated in this paper. Bold indicates the best accuracy per column.}
\resizebox{0.9\textwidth}{!}{%
\begin{tabular}{|c||c|c|c||c|c|c|}

\multicolumn{1}{c}{} & \multicolumn{6}{c}{\textbf{clipViTB16}} \\ \cline{2-7}

\multicolumn{1}{c||}{} & \multicolumn{3}{c||}{\textbf{FACE}} & \multicolumn{3}{c|}{\textbf{PERIOCULAR}} \\ 
 \cline{2-7}

\multicolumn{1}{c||}{\textbf{Features}} & Male & Female & Both & Male & Female & Both \\  \hline

sim (zero shot) & \textbf{95.11} $\pm$ 1.19 & 93.1 $\pm$ 2.6 & 94.06 $\pm$ 1.32 & \textbf{97.7} $\pm$ 0.87 & 49.76 $\pm$ 1.21 & 72.36 $\pm$ 1.61 \\ \hline \hline

sim (aligned) & 93.76 $\pm$ 2.53 & 93.99 $\pm$ 2.63 & 93.95 $\pm$ 1.28 & 81.59 $\pm$ 6.96 & 77.75 $\pm$ 2.02 & 79.66 $\pm$ 3.48 \\ \hline \hline

sim + SVM & 94.37 $\pm$ 1.87 & 94 $\pm$ 2 & \textbf{94.21} $\pm$ 1.37 & 79.55 $\pm$ 5.59 & 80.12 $\pm$ 2.36 & 79.9 $\pm$ 3.35 \\ \hline

\textbf{I} + SVM & 94.13 $\pm$ 1.84 & 93.83 $\pm$ 2.17 & 94.01 $\pm$ 1.41 & 76.67 $\pm$ 4.55 & 82.69 $\pm$ 2.91 & 79.9 $\pm$ 3.21  \\ \hline

(sim + \textbf{I}) + SVM & 94.15 $\pm$ 1.77 & \textbf{94.02} $\pm$ 2.4 & 94.11 $\pm$ 1.56 & 76.96 $\pm$ 4.66 & \textbf{82.84} $\pm$ 2.75 & \textbf{80.12} $\pm$ 3.16  \\ \hline

 \multicolumn{7}{c}{}  \\

\multicolumn{1}{c}{} & \multicolumn{6}{c}{\textbf{clipViTL14}} \\ \cline{2-7}

\multicolumn{1}{c||}{} & \multicolumn{3}{c||}{\textbf{FACE}} & \multicolumn{3}{c|}{\textbf{PERIOCULAR}} \\ 
 \cline{2-7}

\multicolumn{1}{c||}{\textbf{Features}} & Male & Female & Both & Male & Female & Both \\  \hline

sim (zero shot) & \textbf{97.55} $\pm$ 1 & 93.71 $\pm$ 2.2 & 95.54 $\pm$ 1.52 & \textbf{98.8} $\pm$ 0.66 & 54.86 $\pm$ 2.55 & 75.57 $\pm$ 2.07 \\ \hline \hline

sim (aligned) & 95.58 $\pm$ 2.05 & \textbf{96.08} $\pm$ 1.48 & \textbf{95.87} $\pm$ 1.58 & 82.5 $\pm$ 7.85 & \textbf{87.64} $\pm$ 1.42 & 85.29 $\pm$ 4.13 \\ \hline \hline

sim + SVM & 96.17 $\pm$ 1.76 & 95.54 $\pm$ 1.97 & \textbf{95.87} $\pm$ 1.63 & 82.89 $\pm$ 7.55 & 87.46 $\pm$ 1.82 & 85.38 $\pm$ 4.2  \\ \hline

\textbf{I} + SVM & 95.92 $\pm$ 2.17 & 95.58 $\pm$ 1.58 & 95.77 $\pm$ 1.64 & 84.87 $\pm$ 3.86 & 86.67 $\pm$ 2.76 & 85.87 $\pm$ 2.46 \\ \hline

(sim + \textbf{I}) + SVM & 95.83 $\pm$ 1.86 & 95.8 $\pm$ 1.94 & 95.85 $\pm$ 1.68 & 85.08 $\pm$ 3.8 & 87.06 $\pm$ 2.76 & \textbf{86.17} $\pm$ 2.42 \\ \hline

 \multicolumn{7}{c}{}  \\

 \multicolumn{1}{c}{} & \multicolumn{6}{c}{\textbf{clipR50}} \\ \cline{2-7}

\multicolumn{1}{c||}{} & \multicolumn{3}{c||}{\textbf{FACE}} & \multicolumn{3}{c|}{\textbf{PERIOCULAR}} \\ 
 \cline{2-7}

\multicolumn{1}{c||}{\textbf{Features}} & Male & Female & Both & Male & Female & Both \\  \hline

sim (zero shot) & \textbf{94.85} $\pm$ 2.01 & 89.71 $\pm$ 2.6 & 92.17 $\pm$ 1.28 & \textbf{98.97} $\pm$ 0.47 & 33.92 $\pm$ 1 & 64.59 $\pm$ 1.92 \\ \hline \hline

sim (aligned) & 91.71 $\pm$ 4.24 & 92.48 $\pm$ 2.58 & 92.21 $\pm$ 1.53 & 75.93 $\pm$ 6.48 & 76.67 $\pm$ 2.18 & 76.39 $\pm$ 3.24 \\ \hline \hline

sim + SVM & 91.99 $\pm$ 2.66 & 93.4 $\pm$ 1.9 & 92.78 $\pm$ 1.36 & 72.36 $\pm$ 6.75 & \textbf{80.59} $\pm$ 1.73 & \textbf{76.79} $\pm$ 3.37 \\ \hline

\textbf{I} + SVM & 91.58 $\pm$ 3.37 & 93.18 $\pm$ 2.18 & 92.48 $\pm$ 1.67 & 72.67 $\pm$ 5.33 & 79.29 $\pm$ 3.25 & 76.25 $\pm$ 2.75  \\ \hline

(sim + \textbf{I}) + SVM & 92.03 $\pm$ 2.63 & \textbf{93.43} $\pm$ 1.99 & \textbf{92.82} $\pm$ 1.44 & 73.42 $\pm$ 5.16 & 79.37 $\pm$ 3.53 & 76.65 $\pm$ 2.74 \\ \hline

 \multicolumn{7}{c}{}  \\

\end{tabular}
}
\label{tab:results_all}
\end{table*}

\begin{table*}[h]
\centering
\caption{Summary of the best reported accuracy of the experiments of this paper. The SVM combinations shown are those with the best overall accuracy in Table~\ref{tab:results_all}. The table also includes results of recent works using the Adience database.
Different works may not employ the same number of images per fold, so results may not be completely comparable.}
\resizebox{0.9\textwidth}{!}{%
\begin{tabular}{|c|c|c|c|c|}

\multicolumn{5}{c}{\textbf{FACE}} \\ 
\hline 

\textbf{Ref} & \textbf{Features} & \textbf{Male} & \textbf{Female} & \textbf{Both} \\  \hline

this paper: & clipViTB16 & 95.11 $\pm$ 1.19 & 93.1 $\pm$ 2.6 & 94.06 $\pm$ 1.32 \\ \cline{2-5}

sim zero shot  & clipViTL14 & \textbf{97.55} $\pm$ 1 & 93.71 $\pm$ 2.2 & 95.54 $\pm$ 1.52 \\ \cline{2-5} 

  & clipR50 & 94.85 $\pm$ 2.01 & 89.71 $\pm$ 2.6 & 92.17 $\pm$ 1.28 \\ \hline \hline

this paper: & clipViTB16 & 93.76 $\pm$ 2.53 & 93.99 $\pm$ 2.63 & 93.95 $\pm$ 1.28 \\ \cline{2-5}

sim aligned  & clipViTL14 & 95.58 $\pm$ 2.05 & \textbf{96.08} $\pm$ 1.48 & \textbf{95.87} $\pm$ 1.58 \\ \cline{2-5}

  & clipR50 & 91.71 $\pm$ 4.24 & 92.48 $\pm$ 2.58 & 92.21 $\pm$ 1.53 \\ \hline \hline

this paper: & clipViTB16 & 94.37 $\pm$ 1.87 & 94 $\pm$ 2 & 94.21 $\pm$ 1.37 \\ \cline{2-5}

SVM  & clipViTL14 & 96.17 $\pm$ 1.76 & 95.54 $\pm$ 1.97 & \textbf{95.87} $\pm$ 1.63 \\ \cline{2-5} 

  & clipR50 & 92.03 $\pm$ 2.63 & 93.43 $\pm$ 1.99 & 92.82 $\pm$ 1.44 \\ \hline \hline

\cite{[Bhattacharyya19scGenderFacialRegionsGA]} (2019) & compass LBP + SVM &  88.69 & 86.80 & 87.71 \\ \hline 

\cite{[Fang19neurocomputingGenderAgeMultiStage]} (2019) & VGG-19 with local/global relationship &  - & - & 93.52  \\ \hline

\cite{2019_HungICMR_MultipleFaceInformaticsModules} (2019) & ProgressiveNet + PackNet &  - & - & 89.08 \\ \hline 

\cite{2020_KhanSensors_Multitask_Face_Attributes_CNNs} (2020) & RAG-MCFP-DCNNs & - & - & 93.6 \\ \hline 

\cite{[Alonso21ietSoftBio]} (2021) & MobileNetv2 & 84.1 & 86.1 & 85.3 $\pm$ 5.4 \\ \hline  

\cite{2021_BenkaddourInformatica_CNNAgeGender} (2021) & Deep CNN & - & - & 95.6 \\ \hline 

\cite{2023_KuprashevichAIST_MiVOLOAgeGender} (2023) & ViT & - & - & 96.51 \\ \hline 

\cite{2023_GhrbanJCSSP_gender_face_eyes} (2023) & Wavelet Scattering CNN &  - & - & 91 \\ \hline

\cite{2024_SinghNCA_HybridTransformerSequencer} (2024) & ViT + BiLSTM &  - & - & 96.56 $\pm$ 0.27 \\ \hline

\cite{2024_WarisMS_GenderClassification} (2024) & Deep CNN + Soft voting
classifier &  - & - & \textbf{97.61} \\ \hline



 \multicolumn{5}{c}{}  \\

\multicolumn{5}{c}{\textbf{PERIOCULAR (BOTH EYES)}} \\ \hline 

\textbf{Ref} & \textbf{Features} & \textbf{Male} & \textbf{Female} & \textbf{Both} \\  \hline

this paper: & clipViTB16 & 97.7 $\pm$ 0.87 & 49.76 $\pm$ 1.21 & 72.36 $\pm$ 1.61 \\ \cline{2-5}

sim zero shot  & clipViTL14 & 98.8 $\pm$ 0.66 & 54.86 $\pm$ 2.55 & 75.57 $\pm$ 2.07 \\ \cline{2-5} 

  & clipR50 & \textbf{98.97} $\pm$ 0.47 & 33.92 $\pm$ 1 & 64.59 $\pm$ 1.92 \\ \hline \hline

this paper: & clipViTB16 & 81.59 $\pm$ 6.96 & 77.75 $\pm$ 2.02 & 79.66 $\pm$ 3.48 \\ \cline{2-5}

sim aligned  & clipViTL14 & 82.5 $\pm$ 7.85 & \textbf{87.64} $\pm$ 1.42 & 85.29 $\pm$ 4.13 \\ \cline{2-5}

  & clipR50 & 75.93 $\pm$ 6.48 & 76.67 $\pm$ 2.18 & 76.39 $\pm$ 3.24  \\ \hline \hline

this paper:  & clipViTB16 & 76.96 $\pm$ 4.66 & 82.84 $\pm$ 2.75 & 80.12 $\pm$ 3.16 \\ \cline{2-5} 

SVM  & clipViTL14 & 85.08 $\pm$ 3.8 & 87.06 $\pm$ 2.76 & \textbf{86.17} $\pm$ 2.42 \\ \cline{2-5}

  & clipR50 & 72.36 $\pm$ 6.75 & 80.59 $\pm$ 1.73 & 76.79 $\pm$ 3.37 \\ \hline \hline

\cite{[Bhattacharyya19scGenderFacialRegionsGA]} (2019) & compass LBP + SVM &  - & - & 83.27 \\ \hline 

\cite{[Alonso21ietSoftBio]} (2021) & MobileNetv2 & 77.1 & 80.4 & 78.9 $\pm$ 3.7 \\ \hline

\cite{2023_GhrbanJCSSP_gender_face_eyes} (2023) & Deep CNN &  - & - & 83.34 \\ \hline

 \multicolumn{5}{c}{}  \\

\end{tabular}
}
\label{tab:results_soa}
\end{table*}

\subsection{SVM-trained face and periocular gender estimation.}

To investigate whether CLIP embeddings contain discriminative information beyond the zero-shot rule, we train linear SVMs using three feature configurations: $i$) the ten prompt-specific similarity scores; $ii$) the image embeddings; and $iii$) the concatenation of $i$ and $ii$. 
For periocular images, similarity scores use the average of $\mathbf{I_L}$ and $\mathbf{I_R}$, whereas the image-only configuration concatenates both embeddings. 
Thus, configuration $i$ uses the same type of information as the zero-shot approach, $ii$ uses only image encodings, and $iii$ combines both.
Results are summarised in Table~\ref{tab:results_all}.
For full-face estimation, SVMs produce only marginal changes, with per-class differences generally below 0.5\%, and no feature configuration consistently outperforms the others. 
Overall, SVMs match or slightly improve the aligned zero-shot results. 
In addition, the relatively similar performance seen when using image embeddings alone indicates that the text encoder can be discarded when supervised training is allowed.
For periocular estimation, SVMs generally provide slightly higher overall accuracy than threshold alignment. 
Improvements in female accuracy are typically accompanied by decreases in male accuracy, so the main effect is a redistribution of class performance rather than a large overall gain. 
As with full-face, image embeddings alone give competitive results, making the text encoder unnecessary in the supervised setting.

\subsection{Comparison with the state-of-the-art.}

Table~\ref{tab:results_soa} compares our results with previous work on Adience. 
For full-face gender estimation, CLIP achieves competitive performance, with a best accuracy of 95.87\%, within 2\% of the state of the art. 
Notably, the original zero-shot approach already reaches 95.54\% without task-specific training, relying only on image-text similarity, whereas previous methods are predominantly supervised.
For periocular estimation, our best SVM result reaches 86.17\%, outperforming the previous best result of 83.34\% by approximately 2.8\%. 
The aligned approach achieves a comparable 85.29\% using only a calibrated decision threshold, showing that alignment recovers most of the benefit of supervised SVM training. 
Nevertheless, periocular performance remains substantially below full-face results, reflecting the greater difficulty of estimating gender from the smaller facial region.
%


\section{Conclusion}
\label{sec:conclusion}

We investigated CLIP \cite{2021_RadfordICML_CLIP} for gender estimation from full-face and periocular images using zero-shot image-text similarity with two extensions: decision-threshold alignment and linear SVM classifiers. 
Experiments on 11,299 Adience smartphone images showed that CLIP performs competitively on full-face images even without task-specific training, indicating good alignment between image embeddings and gender-related prompts.
For periocular images, however, the original zero-shot formulation showed a strong male bias. 
Score analysis revealed that this behaviour is not explained solely by lower image resolution of periocular crops, but mainly by a shifted decision boundary and greater overlap between male and female score distributions. 
Threshold alignment substantially reduced bias and helped balance class performance, while SVM training provided only modest additional gains. Image embeddings alone were also sufficient to achieve competitive SVM performance, allowing the text encoder to be discarded in the supervised setting.
Comparison with previous works shows that the proposed CLIP-based approach is competitive for full-face gender estimation (95.87\% accuracy, within 2\% of the state of the art), whereas for periocular gender estimation, we outperform previous works on Adience by 2.8\%, achieving 86.17\% accuracy.
Nevertheless, the remaining gap between full-face and periocular gender estimation suggests a weaker separability of male and female periocular representations in the employed CLIP embedding space. 
Future work will investigate improved alignment and prompt-optimisation strategies \cite{2025_OtroshiTIFS_Foundation_Models_Biometrics_Survey}, together with model-compression techniques \cite{caldeira2025inffus_model_compression_biometrics}, including pruning \cite{alonso25PRLdeepnetworkpruning}, knowledge distillation \cite{2024_BoutrosECCV_AdaDistill_KD_FR}, lightweight encoders, and intermediate-layer embeddings \cite{Alonso24wifs_cnn_vit_ots}, to make CLIP-based soft-biometric estimation more practical for deployment.

\section*{Acknowledgment}

This work was partly done while F. A.-F. was a visiting researcher at the University of the Balearic Islands.
F. A.-F., K. H.-D., and J. B. thank the
Swedish Research Council (VR) and the EU (HORIZON Europe project PopEye under Grant Agreement no 101168317). 
Funded by the European Union. Views and opinions expressed are, however, those of the author(s) only and do not necessarily reflect those of the European Union or the European Research Executive Agency. Neither the European Union nor the granting authority can be held responsible for them.
This work is part of the Project PID2022-136779OB-C32 (PLEISAR) funded by MICIU/ AEI /10.13039/501100011033/ and FEDER, EU.

\bibliographystyle{IEEEtran}
\bibliography{fernando1_for_biosig26}

\end{document}